\documentclass{article}
\usepackage{ijcai26}

\usepackage{times}
\usepackage{soul}
\usepackage{url}
\usepackage[hidelinks]{hyperref}
\usepackage[utf8]{inputenc}
\usepackage[small]{caption}
\usepackage{graphicx}
\usepackage{amsmath}
\usepackage{amssymb}
\usepackage{amsthm}
\usepackage{booktabs}
\usepackage{algorithm}
\usepackage{algorithmic}
\usepackage[switch]{lineno}

\title{\textbf{FedLLM-Align}: Feature Extraction From Heterogeneous Clients}

\author{
Abdelrhman Gaber, Muhammad ElMahdy, Youssif Abuzied, Hassan Abd-Eltawab, Tamer ElBatt\\
Computer Science and Engineering Dept.\\
American University in Cairo\\
Cairo, Egypt\\
\texttt{\{gaberabdo68, muhammadahmedelmahdy, youssif.abuzied, hassan.abdeltawab, tamer.elbatt\}@aucegypt.edu}
}

\begin{document}

\maketitle

\begin{abstract}
Federated learning (FL) enables collaborative model training without sharing raw data, making it attractive for privacy-sensitive domains, e.g., healthcare, finance, and IoT. A major obstacle, however, is the potential heterogeneity of tabular data across clients, in practical settings, where schema mismatches and incompatible feature spaces prevent straightforward aggregation. To address this challenge, this paper proposes FedLLM-Align, a federated learning framework that leverages pretrained transformer based language models for feature extraction. Towards this objective, FedLLM-Align serializes tabular records into text and derives semantically aligned embeddings from a pretrained LLM encoder, e.g, DistilBERT, facilitating lightweight local classifier heads that can be trained in a federated manner using standard aggregation schemes, e.g., FedAvg, while keeping all raw data records local. To quantify the merits and trade-offs of FedLLM-Align, we evaluate the proposed framework on binary classification tasks from two different domains: i) Coronary heart disease prediction on partitioned Framingham Heart Study data, and ii) Customer churn prediction on a financial dataset. FedLLM-Align outperforms state-of-the-art baselines by up to 25\% in terms of the F1 score, under simulated schema heterogeneity, and achieves a 65\% reduction in the communication overhead. These results establish FedLLM-Align as a privacy-preserving and communication-efficient approach for federated training based on clients with heterogeneous tabular datasets, commonly encountered in practice.

\end{abstract}
\section{Introduction}

\label{sec:introduction}
Federated learning (FL) enables multiple clients to collaboratively train a global model while keeping their individual training data local. Instead of sharing their individual raw data, FL relies on exchanging model updates (e.g., gradients) between the clients and a central server. This privacy-preserving framework has seen growing adoption in domains with sensitive data, e.g., healthcare, financial systems, and  IoT \cite{yuan2021generalization}. FL is particularly well suited for IoT and edge systems, where devices collect data but regulatory and/or operational constraints, e.g., bandwidth, hinder uploading the raw data \cite{dritsas2025federated} to the cloud. By moving computation to the cloud, FL mitigates privacy and regulatory risks (e.g., the European Union General Data Protection Regulation (GDPR)) \cite{horvath2021fjord} while still enabling global model improvements.

A major hurdle for practical FL deployment is {\em data heterogeneity} across clients. In real-world settings, clients often  

\noindent
collect different attributes, leading to heterogeneous data distributions. For example, user behavior models may observe different features or label distributions on each device. FL must also contend with \emph{system heterogeneity}, where clients differ in hardware and connectivity, and \emph{structural heterogeneity}, where feature spaces or data schemas differ across clients. In clinical settings, electronic health record (EHR) systems at different hospitals may record different attributes or units, a problem known as ``data view heterogeneity''. Such heterogeneity poses a major challenge for FL, potentially degrading performance or even preventing convergence \cite{thakur2024knowledge}.

To address client heterogeneity, prior work has explored several approaches. For instance, {\em personalized FL} (PFL) tailors models to each client’s data \cite{5}, by fine-tuning a global model locally or learning an additional personal model. However, most PFL models primarily address statistical non-IID data and do not account for system or data structural differences, often sacrificing global performance \cite{tan2022personalized}. Another approach is {\em clustered FL}, which groups clients with similar data distributions and trains a separate model per cluster \cite{sattler2020clustered}. Other work proposes knowledge-distillation or transfer methods (e.g., sharing predictions on proxy data) \cite{8} and feature-alignment techniques that map raw inputs into a common latent space \cite{9}. For instance, recent work introduces a ``knowledge abstraction'' mechanism to unify heterogeneous EHR views \cite{gou2021knowledge}. While these methods mitigate data heterogeneity, they present limitations. First, PFL could reduce global generalization, distillation-based schemes often require auxiliary data and may raise privacy concerns, and ensemble approaches can be computationally expensive  \cite{5}. 

In this work, we propose a new direction by leveraging large language models (LLMs) as a feature extraction mechanism to possibly heterogeneous tabular client data before federated training. Recent advances show that LLMs pre-trained on large and diverse corpora can generate strong latent representations for structured data \cite{gardner2024large}. For example, TABULA-8B fine-tunes a Llama-3 8B model on billions of tabular records and achieves a strong zero- and few-shot performance across unseen tasks \cite{gardner2024large}. Inspired by this, we employ LLMs to map each client’s raw tabular features into a shared embedding space. Since the LLM encoder has been exposed to diverse data, its output vectors serve as a common representation, effectively transforming heterogeneous client data into homogeneous embeddings suitable for downstream FL models.

The main contribution of this work is multifold. First, we introduce a novel federated learning framework in which a pre-trained LLM acts as a client-agnostic feature encoder for tabular data, and illustrate how to tokenize and encode client-specific records to produce consistent embeddings. Second, we quantitatively show that training on these embeddings significantly improves cross-client performance under data heterogeneity compared to baseline FL. Finally, we evaluate the proposed framework across diverse tasks and heterogeneity settings, demonstrating its advantages over the personalization and clustering baselines. 

The rest of this paper is organized as follows. Section~\ref{sec:related} surveys the related literature. Section~\ref{sec:method} presents the proposed LLM-based encoding approach for handling structured data heterogeneity in federated learning. Experimental results and a discussion are provided in Sections 4, 5, and 6. Finally, Section~\ref{sec:conclusion} concludes the paper and outlines future research directions.

\section{Related Work}
\label{sec:related}
There has been increasing interest in using large language models (LLMs) to address data heterogeneity in tabular learning. For example, TabLLM~\cite{hegselmann2023tabllm} introduces a few-shot tabular classification method by converting rows into natural-language strings and prompting LLMs (e.g., T0, GPT-3). It explores various serialization strategies and uses parameter-efficient fine-tuning (T-Few) to adapt the LLM. TabLLM demonstrates strong zero- and few-shot performance, often surpassing gradient-boosted trees and neural baselines. Its benefits include sample efficiency and leveraging prior LLM knowledge, while limitations involve high computational cost, token limits, and reliance on semantically meaningful features. More recently, Latte~\cite{ijcai2025p687} showed that transferring latent-level knowledge from pretrained LLMs further improves few-shot tabular learning, emphasizing the value of LLM representations over purely text-level features.

Another approach, FeatLLM~\cite{han2024llmfeatures}, proposes an in-context learning framework where LLMs serve as feature engineers for few-shot tabular learning. Instead of end-to-end inference, the LLM generates interpretable rules from a few examples, which are then transformed into binary features for lightweight models. Bagging ensembles improve robustness and mitigate prompt size limits. FeatLLM achieves state-of-the-art results across 13 datasets with lower inference cost. Its main advantages are low latency and feature interpretability, while limitations include sensitivity to prompt quality and applicability only in low-shot settings.

Another related approach is PTab~\cite{liu2022ptab}, a three-stage framework for modeling tabular data with pretrained language models. It mitigates semantic gaps by converting rows into text (Modality Transformation), followed by Masked-Language Fine-tuning and Classification Fine-tuning. This textualization bridges domain differences and allows training on mixed tabular datasets. Evaluated on eight binary classification tasks, PTab outperforms XGBoost and neural baselines (e.g., SAINT, TabTransformer) in average AUC under both supervised and semi-supervised settings.

Closest to our federated setting is SecEA (Secure Embedding Aggregation)~\cite{tang2022privacy}, which introduces a secure embedding aggregation protocol for federated representation learning, providing information-theoretic privacy against a curious server and up to $T < N/2$ colluding clients. SecEA performs a private entity union and distributes local embeddings via secret sharing and Lagrange coded computing. Across tasks like knowledge graph completion, recommendation, and node classification, SecEA incurs under $5\%$ performance loss compared to non-private baselines while achieving notable efficiency gains through parallelization. Complementing this, recent IJCAI work, such as CReFF~\cite{ijcai2022p308} studies federated learning on heterogeneous data by decoupling representation learning and classifier re-training, showing that carefully designed representation and aggregation schemes are key for robust FL under non-IID client distributions.

Taken together, the approaches reviewed above (e.g., TabLLM-/Latte-style LLM-based tabular learning, FeatLLM-style LLM feature engineering, PTab-style textualization frameworks, and SecEA/CReFF-style federated representation and aggregation methods) often assume a globally aligned feature space, rely on centrally curated or mixed datasets, or primarily focus on privacy and statistical heterogeneity without explicitly addressing schema-level heterogeneity across clients. Moreover, many require exchanging full model parameters, gradients, or high-dimensional embeddings, leading to high communication overhead. In contrast, FedLLM-Align is specifically designed for federated learning over heterogeneous tabular schemas, using LLM-based universal encoders to align client feature spaces while maintaining low communication cost.

\section{Proposed Methodology: FedLLM-Align} \label{sec:method}
This section presents the proposed FedLLM-Align framework, which addresses schema-level heterogeneity in federated learning through LLM-based semantic feature alignment. We first describe the federated learning problem under heterogeneous tabular schemas, then describe the architecture and training pipeline of FedLLM-Align. The methodology details how tabular records are serialized, embedded using frozen pretrained language models, and integrated into standard federated optimization schemes while preserving data privacy and communication efficiency.
\subsection{System Model}
We consider a set of $N$ clients, denoted $U_1, U_2, \ldots, U_N$, with private datasets, denoted $D_1, D_2, \ldots, D_N$. The local dataset at client $U_i$, $D_i$, hosts structured data in the form of tabular records defined over a schema of features, $S_i = \{f_1^i, f_2^i, \ldots, f_{m_i}^i\}$, where feature names and representations may differ across clients. The problem setting imposes a number of constraints. First, schema heterogeneity is assumed inherent in practical settings, since the overlap between two schemas, e.g., $S_i$ and $S_j$, may be small or even empty for any $i \neq j$. Second, the proposed solution must remain compatible with standard federated aggregation schemes, such as FedAvg, so that it can be seamlessly integrated into existing federated learning pipelines. 

\subsection{FedLLM-Align Architecture}
\begin{figure}[h]
\centering
\includegraphics[width=\linewidth, height=0.24\textwidth]{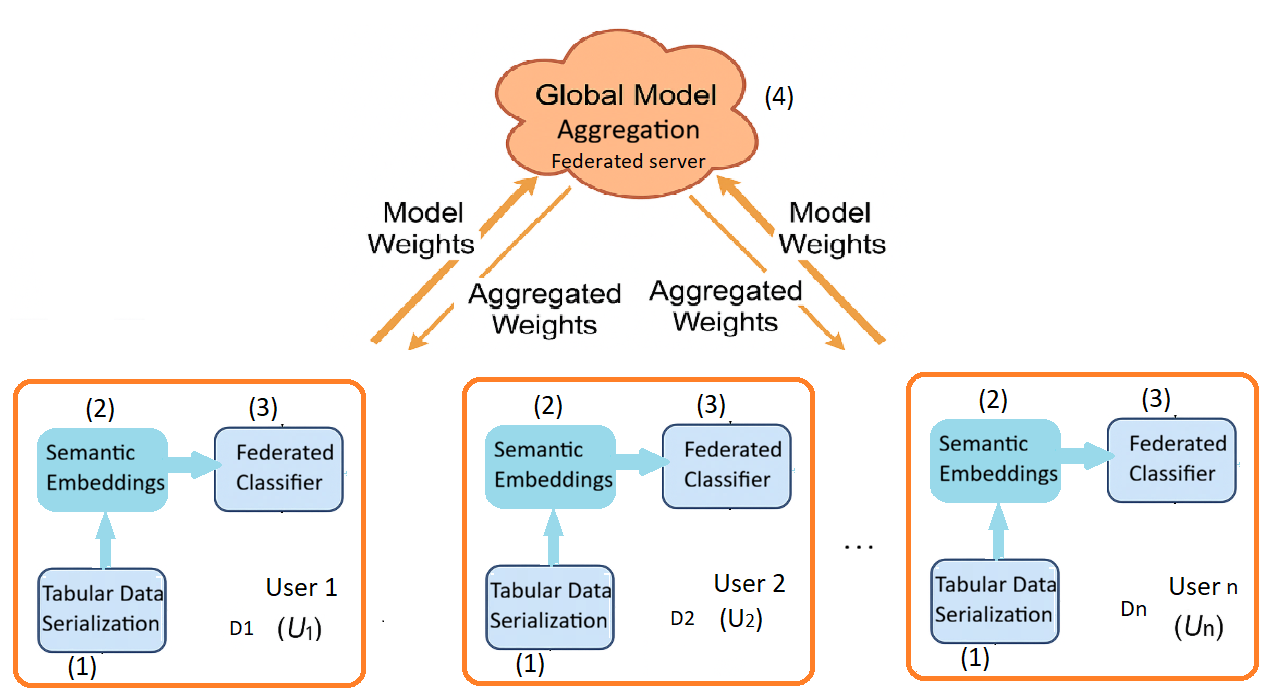}
\caption{FedLLM-Align Pipeline: (1) Tabular-to-text conversion, (2) Embedding generation, (3) On-device classifier training, (4) Global weight aggregation using FedAvg.}
\label{fig:architecture}
\end{figure}

The FedLLM-Align framework addresses the above challenges through a three-stage pipeline: i) tabular-to-text serialization, ii) LLM feature extraction: semantic embedding generation, and iii) On-device classifier training, in addition to the federated model aggregation, as shown in Figure~\ref{fig:architecture}. Next, we introduce the proposed three-stage pipeline for LLM feature extraction, its technical rationale, and design trade-offs. 

\subsubsection{I. Tabular-to-Text Serialization}
At the first stage, each client, $i$, transforms its local records $\mathbf{x}_j \in D_i$ into natural language sequences through a serialization function,

\begin{equation}
    \text{serialize}(\mathbf{x}_j, \text{format}) \rightarrow \text{text\_sequence}.
\end{equation}

Different serialization strategies may be applied. A structured format explicitly lists features and values, ``Feature$_1$: value$_1$, Feature$_2$: value$_2$, ...''. A natural language format encodes features in descriptive sentences, such as ``The patient is 45 years old. Blood pressure is 140/90.''. A compact format, instead, may use condensed key-value pairs, ``Feature$_1$=value$_1$; Feature$_2$=value$_2$; ...''. The key insight of FedLLM-Align is that by serializing tabular records into short, structured natural-language descriptions, pretrained LLMs can exploit their semantic understanding to align semantically equivalent features across heterogeneous client schemas. In our main experiments we adopt this structured serialization, and later compare it against more free-form and compact variants.

\subsubsection{II. Semantic Embedding Generation}
At the second stage, each serialized sequence is passed through a frozen pretrained LLM, producing a semantic embedding,
\vspace{-3mm}
\begin{equation}
    \mathbf{e}_j = \text{LLM\_encoder}(\text{text\_sequence})_{[\text{CLS}]} \in \mathbb{R}^d.
\end{equation}
\vspace{-3mm}

Among the supported backbones, DistilBERT provides a lightweight six-layer distilled BERT model that balances efficiency with representational quality, while ALBERT\cite{lan2019albert} applies parameter sharing to achieve memory efficiency with competitive embedding quality. It is worth noting that these LLM backbones remain frozen during the training process. This design choice reduces the communications overhead, e.g., over bandwidth-limited wireless links, by ensuring that only the classifier weights are exchanged, preserves the pretrained semantic knowledge of the models, and supports deployment across clients with limited computational resources at the network edge.

\subsubsection{III. The Federated Classifier}
In the final stage, the generated embeddings $\{\mathbf{e}_1, \mathbf{e}_2, \ldots, \mathbf{e}_m\}$, each in $\mathbb{R}^d$, serve as input to a lightweight classifier trained locally on each client, where $d$ denotes the dimensionality of the encoder’s output representation. As a proof of concept, we adopt a single downstream model in the federated setting: a shallow feedforward neural network that operates directly on the frozen LLM embeddings and produces a scalar output for binary prediction. During federated training, only the classifier parameters are shared with the central server, while both the LLM-based embeddings and the raw tabular records remain strictly local. This design preserves data locality and offers a practical security advantage, since the global model operates over a shared semantic embedding space rather than client-specific raw feature schemas.

\subsection{Federated Training}
The training pipeline is summarized in Algorithm~\ref{alg:fedllm_align}.
\begin{algorithm}[H]
\caption{FedLLM-Align Training Pipeline}
\label{alg:fedllm_align}

\textbf{Input:} Client Datasets $D_1, \ldots, D_N$, with heterogeneous schemas

\textbf{Ensure:} Global Classifier Model $M_{\text{global}}$

\begin{algorithmic}[1]
\STATE Initialize weights $W_0$ of $M_{\text{global}}$
\FOR{ round $t = 1$ to $T$}
    \STATE Sample subset $S_t \subseteq \{1, \ldots, N\}$
    \FOR{each client $i \in S_t$ \textbf{in parallel}}
        \STATE Perform tabular-to-text serialization for each record $\mathbf{x}_j \in D_i$
        \STATE Compute embeddings $\mathbf{e}_j$ using frozen LLM
        \STATE Train local classifier $M_i$ with initialized weights $W^t$
        \STATE Send weight updates $\Delta W_i = W_i^{t+1} - W^t$ to server
    \ENDFOR
    \STATE Aggregate updates: $W^{t+1} = W^t + \frac{1}{|S_t|} \sum_{i \in S_t} \Delta W_i$
    \STATE Broadcast updated weights $W^{t+1}$ to all clients
\ENDFOR
\end{algorithmic}
\end{algorithm}
In each federated training round, a subset of clients is selected to participate and initialize their local classifiers with the current global model parameters. Each participating client first serializes its local tabular records into textual representations and computes semantic embeddings using a frozen pretrained language model. These embeddings are then used to train a lightweight classifier locally for several epochs. Upon completion, only the resulting classifier weight updates are transmitted to the central server, where they are aggregated using a standard parameter-averaging rule such as FedAvg to update the global model. The updated parameters are subsequently broadcast back to the clients for the next round, and this process is repeated until convergence.
\subsection{FedLLM-Align Merits}
The proposed framework provides three operational guarantees. First, \emph{semantic alignment} arises from the pre-trained LLMs, which map semantically equivalent attributes and values to nearby points in the embedding space, even when feature labels differ. For example, ``Age: 45'' and ``PatientAge: 45 years'' yield similar embeddings, as do ``BP: 140/90'' and ``BloodPressure: systolic=140, diastolic=90''. Second, \emph{privacy preservation} is ensured as raw tabular data and intermediate embeddings never leave client devices; only lightweight classifier parameters are sent to the server. Finally, since the encoder is frozen and induces a fixed feature space, \emph{training convergence} follows standard results for parameter-averaging federated optimizers: aggregation rules such as FedAvg can be applied to the classifier layer without modification and retain their usual convergence properties under standard smoothness and bounded-variance assumptions.

\section{Performance Evaluation}
\label{sec:experiments}

In this section, we describe the experimental setup used to evaluate the proposed FedLLM-Align framework. We first outline the baseline schemes, then present the used datasets and how schema heterogeneity is simulated. We next describe the data and feature processing pipeline, and finally introduce the evaluation metrics used in the experiments.

\subsection{Baseline Schemes}
We compare our proposed framework with both traditional and advanced federated learning approaches. Traditional FL models include FedXGBoost~\cite{le2021fedxgboost}, Mutual Information-based FL~\cite{uddin2020mutual}, FedProx~\cite{li2020fedprox} and SCAFFOLD~\cite{karimireddy2020scaffold}. For advanced FL models, we consider two baselines, namely Clustered FL~\cite{sattler2020clustered} and FedAvg with identical schemas and homogeneous tabular data, which serves as a strong reference point.

\subsection{Adopted Datasets}

\subsubsection{Datasets Description}
We evaluate FedLLM-Align on two public datasets from different domains: a financial customer churn dataset \cite{topre2022bankchurn}, and the Framingham Heart Study cardiovascular dataset ~\cite{data}.

\textbf{Financial customer churn (banking).}
The first dataset consists of 10{,}000 retail banking customers, each described by demographic and account-related attributes. The prediction task is to determine whether a customer will exit the bank (churn). The dataset is moderately imbalanced, with approximately 80\% non-churn and 20\% churn samples, similar to realistic bank-attrition rates. The input features include \textit{CreditScore}, \textit{Geography}, \textit{Gender}, \textit{Age}, \textit{Tenure}, \textit{Balance}, \textit{NumOfProducts}, \textit{HasCrCard}, \textit{IsActiveMember}, and \textit{EstimatedSalary}, while the target label \textit{Exited} indicates whether the customer has churned (1) or stayed (0).

\textbf{Framingham Heart Study (healthcare).}
The second dataset is derived from the Framingham Heart Study, a longitudinal cardiovascular cohort of residents in Framingham, Massachusetts, USA. After cleaning, we use 4{,}240 patient records described by 15 demographic, lifestyle, and clinical attributes together with a binary target indicating the 10-year risk of coronary heart disease (CHD). Approximately 85\% of the records belong to the negative class (no CHD) and 15\% to the positive class (CHD), which mirrors real-world prevalence rates. The input features include \textit{Sex}, \textit{Age}, \textit{is\_smoking}, \textit{CigsPerDay}, \textit{BPMeds}, \textit{PrevalentStroke}, \textit{PrevalentHyp}, \textit{Diabetes}, \textit{TotChol}, \textit{SysBP}, \textit{DiaBP}, \textit{BMI}, \textit{HeartRate}, and \textit{Glucose}, while the target label \textit{TenYearCHD} indicates whether the patient develops CHD within 10 years (1) or not (0).

\subsubsection{Datasets Preparation and Heterogeneity Simulation}
To emulate real-world schema misalignment, we introduce controlled heterogeneity in two ways: (i) each client observes only a subset of the available features, and (ii) overlapping features are systematically renamed using alternative but semantically equivalent labels. This procedure is applied independently to both the financial and healthcare datasets, reflecting how different institutions may log related quantities under different names or templates.

Table~\ref{tab:renaming} shows examples of alternative naming conventions for representative features from both domains. For each original feature, clients receive independently sampled aliases drawn from these sets when serializing their local tabular records into text. This preserves semantic meaning while breaking syntactic alignment at the schema level.

\begin{table}[ht!]
\centering
\caption{Examples of Schema Heterogeneity via Feature Renaming}
\label{tab:renaming}
\resizebox{\linewidth}{!}{
\begin{tabular}{|c|p{6cm}|} 
\hline
\textbf{Original Feature} & \textbf{Alternative Names} \\
\hline
\textbf{Age (Framingham)} & \shortstack{Age, PatientAge, AgeYears,\\ age\_at\_visit, patient\_age\_years} \\
\hline
\textbf{SysBP} & \shortstack{SysBP, systolic\_bp, bp\_systolic,\\ sys\_blood\_pressure, systolic\_pressure} \\
\hline
\textbf{TotChol} & \shortstack{TotChol, total\_cholesterol, cholesterol\_total,\\ chol\_total, total\_chol\_mg} \\
\hline
\textbf{CreditScore} & \shortstack{CreditScore, credit\_score, risk\_score,\\ customer\_credit\_rating, credit\_index} \\
\hline
\textbf{Balance} & \shortstack{Balance, account\_balance, cur\_balance,\\ dep\_balance, current\_account\_value} \\
\hline
\textbf{EstimatedSalary} & \shortstack{EstimatedSalary, salary\_est, annual\_income,\\ income\_estimate, yearly\_salary} \\
\hline
\end{tabular}
}
\end{table}

Clients are configured under three federated scenarios corresponding to different collaboration levels:
\begin{itemize}
    \item 3 clients with an overlap ratio of approximately 60\% shared features and the remaining features partitioned into client-specific subsets,
    \item 5 clients with an overlap ratio of approximately 50\% shared features,
    \item 10 clients with an overlap ratio of approximately 40\% shared features.
\end{itemize}

In all cases, the exact subsets are sampled at random, given the desired overlap ratio, ensuring structural heterogeneity (due to partial feature visibility and schema variations). This setup closely reflects cross-institutional settings in healthcare and finance.  

\subsection{Data and Feature Processing Pipeline}
The first step in the proposed pipeline in Fig.~\ref{fig:architecture} is tabular-to-text serialization. We first perform basic preprocessing: missing numerical values are imputed with the median, and categorical variables with the mode. Each record is then serialized into one of three textual formats---\emph{structured}, \emph{natural language}, or \emph{compact}---before tokenization. This serialization exposes attribute names and values as short textual phrases, enabling the downstream language model to exploit its semantic prior over feature names and categories.

The second step is embedding generation using pretrained LLMs. We adopt representative transformer-based encoders (e.g., DistilBERT, ALBERT, RoBERTa, ClinicalBERT) as frozen encoders. For each serialized record, we apply the corresponding fast tokenizer (e.g., \texttt{DistilBertTokenizerFast}, \texttt{AlbertTokenizerFast}) with a maximum sequence length of 128 tokens and feed the tokenized text into the LLM. The [CLS] embedding from the final hidden layer represents each record as a dense vector, which is passed to a lightweight feedforward neural network classifier (input dimension 768, one hidden layer with 16 ReLU units, dropout $p=0.2$, sigmoid output). All LLM encoders remain frozen during training, so only the classifier head is updated and communicated, reducing computation and communication overhead. Finally, as a proof-of-concept, we adopt federated learning with {\it FedAvg} over 25 global aggregation rounds, training clients for 10 local epochs per round with a batch size of 32 using the Adam optimizer ($\text{lr}=0.001$).

\subsection{Performance Metrics}
The primary evaluation metric is the F1-score, complemented by paired t-tests ($\alpha = 0.05$) for statistical significance. In addition, we analyze the communication cost, convergence behavior, per-client performance variance, model memory footprint, and inference latency for embedding extraction. These metrics jointly capture both the predictive effectiveness and the system efficiency. Given that the adopted datasets are imbalanced with respect to the positive class (as mentioned earlier), we focus on the F1-score, which balances precision and recall for the positive class and is therefore more informative than accuracy alone.

\section{Experiments Setup and Results}
\label{sec:results}

All experiments were performed in an environment equipped with a T4 GPU and 12 GB of system memory. The software stack included Python, PyTorch, HuggingFace Transformers, TensorFlow, and Scikit-learn. Unless otherwise stated, we use LLM encoders with embedding dimension $d=768$ and a shallow feedforward classifier on top of the frozen embeddings, consisting of one hidden layer with 16 ReLU units and a sigmoid output for binary prediction. Federated training is carried out with FedAvg for 25 global aggregation rounds, using 10 local epochs per round, a batch size of 32, and the Adam optimizer with learning rate $10^{-3}$. We present the experimental results from this setup next, first comparing different LLM encoders in FedLLM-Align, then examining serialization strategies, client scaling, and schema heterogeneity, convergence and stability, communication efficiency, and finally a stress test for schema overlap.

\subsection{LLM Models Comparison}

In this experiment, we fix the data partitioning, the serialization strategy (structured format), the number of clients, and the federated training protocol, and we vary only the underlying LLM encoder used to generate the embeddings. Specifically, we instantiate FedLLM-Align with several pretrained transformer encoders (DistilBERT, ALBERT, RoBERTa, and ClinicalBERT), keeping all downstream classifier and FedAvg hyperparameters identical. This allows us to quantify the accuracy–efficiency trade-offs (F1-score, memory footprint, and per-record inference time) associated with different encoder architectures while holding the rest of the pipeline constant.

Table~\ref{tab:arch_ablation} shows that DistilBERT achieves the best accuracy–efficiency balance, with an F1-score of 0.84 while requiring only 255 MB of memory and 45 msec inference time per record. ALBERT is more memory-efficient (180 MB), yet yields a slightly lower F1-score. ClinicalBERT provides the highest overall accuracy (0.85) owing to its medical domain pretraining, but at a higher computational cost. RoBERTa falls between these extremes. These results suggest that resource-constrained clients may favor DistilBERT or ALBERT, while ClinicalBERT is ideal in settings where maximizing predictive performance is the most important.

\begin{table}[ht!]
\centering
\caption{LLM Models Comparison (F1-Score $\pm$ Std, Memory, and Inference Time). Bold indicates best performance.}
\label{tab:arch_ablation}
\resizebox{\linewidth}{!}{
\begin{tabular}{|l|c|c|c|}
\hline
\textbf{Encoder} & \textbf{F1-Score} & \textbf{Memory (MB)} & \textbf{Inference Time (ms)} \\
\hline
DistilBERT & 0.84$\pm$0.01 & 255 & 45$\pm$5 \\
ALBERT & 0.81$\pm$0.02 & \textbf{180} & \textbf{38$\pm$4} \\
RoBERTa & 0.83$\pm$0.01 & 498 & 72$\pm$8 \\
ClinicalBERT & \textbf{0.85$\pm$0.01} & 440 & 68$\pm$7 \\
\hline
\end{tabular}
}
\end{table}


\subsection{Serialization Scheme Comparison}
Here, we fix the LLM encoder (DistilBERT), client splits, and federated optimization settings, and instead vary how tabular records are converted to text. Each row is serialized using one of three formats: (i) a structured ``key: value'' style, (ii) a more verbose natural-language description, and (iii) a compact, minimally redundant encoding. For each format, we recompute embeddings and re-run federated training, comparing the resulting F1-scores and cross-client variability. This experiment isolates the impact of the tabular-to-text representation on the quality and stability of the learned embeddings.

As shown in Table~\ref{tab:serialization}, structured serialization consistently yields the highest F1-score (0.84) and most stable embeddings, while natural language adds flexibility but with slightly higher variance. Compact formats are the most efficient but perform poorly due to the loss of semantic richness. This highlights that both model choice and data representation strongly affect the proposed FedLLM-Align performance.

\begin{table}[ht!]
\centering
\caption{Serialization Scheme Comparison}
\label{tab:serialization}
\resizebox{\linewidth}{!}{
\begin{tabular}{|l|c|c|c|}
\hline
\textbf{Format} & \textbf{F1-Score} & \textbf{Embedding Variance} & \textbf{Robustness} \\
\hline
Structured & \textbf{0.84$\pm$0.01} & \textbf{0.12} & High \\
Natural & 0.82$\pm$0.02 & 0.18 & Medium \\
Compact & 0.79$\pm$0.03 & 0.25 & Low \\
\hline
\end{tabular}
}
\end{table}

\subsection{FedLLM-Align: A Comparative Analysis}
\subsubsection{Client Scaling Analysis (Two Datasets)}
We next examine how FedLLM-Align and the baselines behave as the number of clients and schema overlap vary. Deployments with 3, 5, and 10 clients are considered, keeping the total dataset size fixed while redistributing records and features to simulate increasing heterogeneity (fewer shared features and more client-specific attributes as client count grows). For each configuration, all methods are trained with the same number of communication rounds, and global F1-scores, per-client statistics, and communication cost are reported. This setup highlights each approach’s robustness to scaling and more fragmented schemas.

Table~\ref{tab:main_results} compares FedLLM-Align with multiple FL baselines on the Framingham cardiovascular risk prediction task across different client configurations. The results show that FedLLM-Align consistently achieves superior F1-scores, with statistically significant improvements ($p < 0.001$). For example, with three clients, FedLLM-Align (DistilBERT + NN) achieves an F1-score of \textbf{0.84}, outperforming the homogeneous baseline (0.64) and FedXGBoost (0.14). As the number of clients increases to ten, FedLLM-Align maintains strong performance (\textbf{0.78} with DistilBERT), while competing approaches degrade under schema heterogeneity. Importantly, these gains are coupled with efficiency: communication cost is reduced by about 65\% compared to FedXGBoost and remains competitive with other baselines.

\begin{table}[ht!]
\centering
\caption{F1-Score Performance Comparison on the Framingham dataset (Mean $\pm$ Std over 5 runs).}
\label{tab:main_results}
\resizebox{\linewidth}{!}{
\begin{tabular}{|l|c|c|c|c|}
\hline
\textbf{Method} &
\textbf{\shortstack{3\\Clients}} &
\textbf{\shortstack{5\\Clients}} &
\textbf{\shortstack{10\\Clients}} &
\textbf{\shortstack{Avg. Comm.\\Cost (MB)}} \\
\hline
\textbf{\shortstack{FedLLM-Align\\(DistilBERT + NN)}} &
\textbf{0.84$\pm$0.01} &
\textbf{0.81$\pm$0.02} &
\textbf{0.78$\pm$0.02} &
\textbf{1.2} \\
\hline
Homogeneous Baseline &
0.64$\pm$0.02 & 0.62$\pm$0.03 & 0.59$\pm$0.03 & 0.9 \\
FedXGBoost &
0.14$\pm$0.02 & 0.11$\pm$0.03 & 0.08$\pm$0.02 & 3.8 \\
Mutual Information FL &
0.61$\pm$0.03 & 0.54$\pm$0.04 & 0.47$\pm$0.05 & 1.1 \\
FedProx &
0.66$\pm$0.02 & 0.61$\pm$0.03 & 0.56$\pm$0.04 & 1.0 \\
SCAFFOLD &
0.68$\pm$0.02 & 0.63$\pm$0.02 & 0.58$\pm$0.03 & 1.1 \\
Clustered FL &
0.59$\pm$0.05 & 0.52$\pm$0.06 & 0.44$\pm$0.07 & 1.6 \\
\hline
\end{tabular}
}
\end{table}

To complement the cardiovascular study, we also evaluate FedLLM-Align in the bank customer churn data set, which represents a different domain (finance) with distinct feature semantics, but similar class imbalance and business constraints. Table~\ref{tab:main_results_churn} reports F1-scores in the same family of methods and client configurations.

\begin{table}[ht!]
\centering
\caption{F1-Score Performance Comparison on the churn dataset (Mean $\pm$ Std over 5 runs).}
\label{tab:main_results_churn}
\resizebox{\linewidth}{!}{
\begin{tabular}{|l|c|c|c|c|}
\hline
\textbf{Method} &
\textbf{\shortstack{3\\Clients}} &
\textbf{\shortstack{5\\Clients}} &
\textbf{\shortstack{10\\Clients}} &
\textbf{\shortstack{Avg. Comm.\\Cost (MB)}} \\
\hline
\textbf{\shortstack{FedLLM-Align\\(DistilBERT + NN)}} &
\textbf{0.80$\pm$0.02} &
\textbf{0.77$\pm$0.02} &
\textbf{0.73$\pm$0.03} &
\textbf{1.3} \\
\hline
Homogeneous Baseline &
0.70$\pm$0.02 & 0.67$\pm$0.03 & 0.62$\pm$0.03 & 1.0 \\
FedXGBoost &
0.62$\pm$0.03 & 0.60$\pm$0.03 & 0.57$\pm$0.04 & 3.9 \\
Mutual Information FL &
0.68$\pm$0.03 & 0.64$\pm$0.04 & 0.59$\pm$0.04 & 1.2 \\
FedProx &
0.71$\pm$0.02 & 0.68$\pm$0.03 & 0.63$\pm$0.03 & 1.0 \\
SCAFFOLD &
0.72$\pm$0.02 & 0.69$\pm$0.03 & 0.64$\pm$0.03 & 1.2 \\
Clustered FL &
0.66$\pm$0.04 & 0.61$\pm$0.05 & 0.55$\pm$0.06 & 1.7 \\
\hline
\end{tabular}
}
\end{table}

FedLLM-Align achieves the best performance in all client counts in the churn task, with DistilBERT-based FedLLM-Align achieving the highest absolute F1-scores and a favorable accuracy–efficiency trade-off. As the number of clients increases from three to ten, all methods experience some degradation, but FedLLM-Align maintains a clear and consistent margin over optimization-focused baselines such as FedProx and SCAFFOLD, and matches or exceeds alignment-based approaches like Mutual Information FL, while retaining substantially lower communication cost than communication-heavy methods such as FedXGBoost and Clustered FL. This cross-domain consistency supports the claim that FedLLM-Align is a robust option for heterogeneous FL in both healthcare and financial applications.

\subsubsection{Convergence Analysis}
To assess the training robustness of FedLLM-Align, we monitor the convergence behavior across federated rounds and evaluate cross-client performance variance. We track F1-scores over 25 communication rounds and compute per-client statistics (mean, standard deviation, minimum, and maximum F1-scores) to measure performance equity across participants.

\begin{figure}[ht!]

\centering
\includegraphics[width=\linewidth, height=0.33\textwidth]{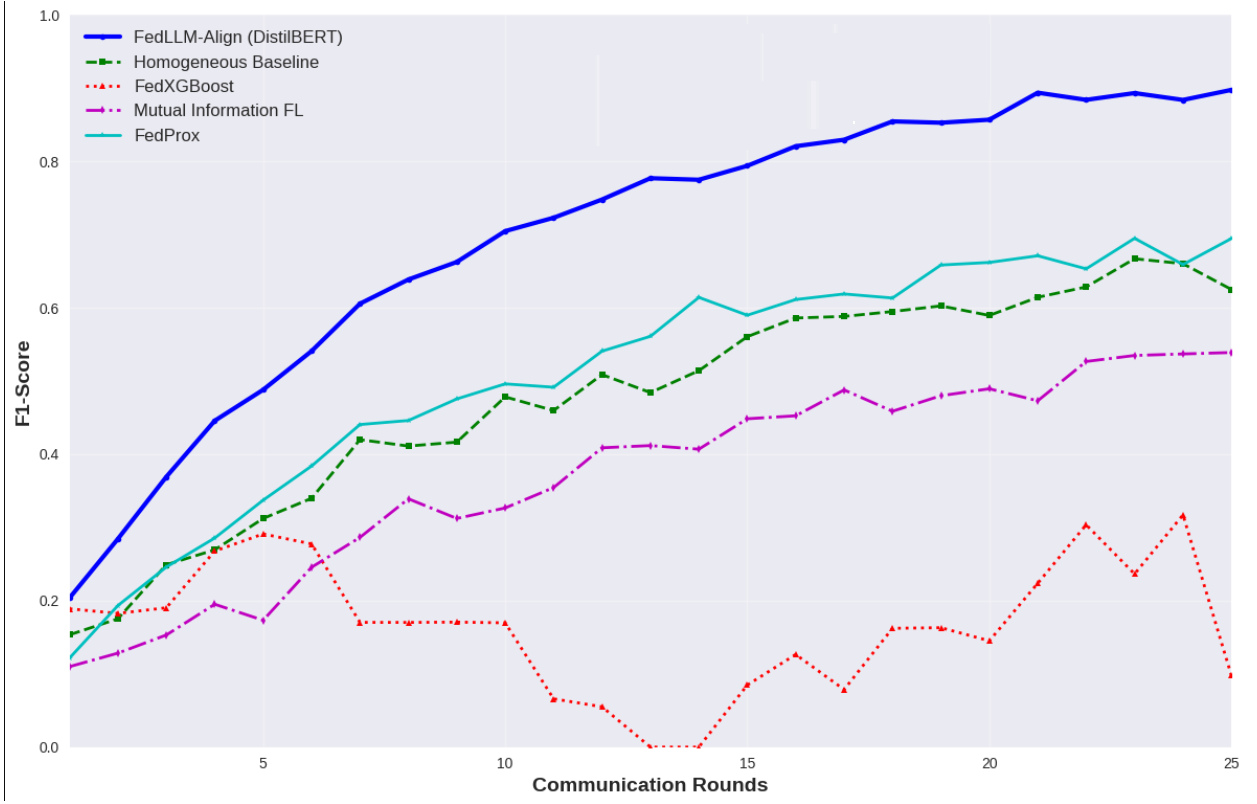}
\caption{Training convergence comparison. FedLLM-Align converges reliably within 15 rounds, unlike baselines.}
\label{fig_convergence_}
\end{figure}

Training dynamics further validate the robustness of FedLLM-Align. Figure~\ref{fig_convergence_} shows that our framework converges smoothly within 15 rounds, whereas FedProx exhibit unstable patterns due to the schema misalignment. Table~\ref{tab:stability} confirms that FedLLM-Align maintains both high accuracy and low cross-client variance and standard deviation (Std = 0.02), ensuring equitable performance across participants. In contrast, FedXGBoost and the homogeneous baseline show wide fluctuations and poor stability, indicating fragile adaptation.

\begin{table}[ht!]
\centering
\caption{Cross-Client Stability (Mean F1, Std, Min, Max). FedLLM-Align shows the lowest variance.}
\label{tab:stability}
\begin{tabular}{|l|c|c|c|c|}
\hline
\textbf{Method} & \textbf{Mean F1} & \textbf{Std} & \textbf{Min F1} & \textbf{Max F1} \\
\hline
FedLLM-Align & \textbf{0.84} & \textbf{0.02} & 0.81 & 0.86 \\
Homogeneous & 0.64 & 0.08 & 0.52 & 0.73 \\
FedXGBoost & 0.14 & 0.12 & 0.02 & 0.31 \\
\hline
\end{tabular}
\end{table}

\subsubsection{Communication Efficiency Analysis}
Communication efficiency is a fundamental requirement in FL since excessive communication overhead can significantly limit scalability and practical deployment. To quantify the communication cost of different methods, we measure the per-round communication overhead, decomposed into model weight transmission and additional protocol-related overhead. We track the size of transmitted updates until convergence for all methods.

Table~\ref{tab:comm_cost} reports the per-round communication overhead. FedLLM-Align achieves the lowest communication cost at 13.1 KB per round, which is 4.1$\times$ lower than FedXGBoost and 1.4$\times$ lower than Mutual Information FL. This efficiency positions FedLLM-Align to be well-suited for bandwidth-constrained federated learning settings while still maintaining competitive performance.

\begin{table}[ht!]
\centering
\caption{Communication Cost Analysis per Round}
\label{tab:comm_cost}
\resizebox{\linewidth}{!}{
\begin{tabular}{|l|c|c|c|c|}
\hline
\textbf{Method} &
\textbf{\shortstack{Model\\Weights (KB)}} &
\textbf{\shortstack{Overhead\\(KB)}} &
\textbf{\shortstack{Total\\(KB)}} &
\textbf{\shortstack{Relative\\Cost}}
\\
\hline
FedLLM-Align & 12.3 & 0.8 & \textbf{13.1} & 1.0$\times$ \\
FedXGBoost & 45.7 & 8.2 & 53.9 & 4.1$\times$ \\
Mutual Info FL & 15.2 & 3.1 & 18.3 & 1.4$\times$ \\
\hline
\end{tabular}
}
\end{table}

\subsubsection{Schema Heterogeneity Stress Test}
Finally, we stress-test the FedLLM-Align framework by progressively reducing the fraction of features that are shared across clients (schema overlap) from 80\% to 20\%. For each overlap level, we re-partition the data and re-run all methods. In this experiment, the \emph{homogeneous} baseline refers to a FedAvg model trained on a \emph{globally aligned schema restricted to the subset of features that is common to all clients at that overlap level}. Thus, unlike in the main results (where homogeneous FedAvg serves as an idealized upper bound assuming fully harmonized schemas), here it is a strong but \emph{overlap-aware} baseline that loses predictive features as the shared subset shrinks.

\begin{table}[ht!]
\centering
\caption{Stress Test Under Schema Divergence. FedLLM-Align degrades gracefully.}
\label{tab:heterogeneity_stress}
\resizebox{\linewidth}{!}{
\begin{tabular}{|l|c|c|c|c|}
\hline
\textbf{\shortstack{Schema\\Overlap}} &
\textbf{\shortstack{FedLLM\\Align}} &
\textbf{\shortstack{Homo-\\geneous}} &
\textbf{\shortstack{Fed\\XGBoost}} &
\textbf{\shortstack{Mutual\\Info}} \\
\hline
80\% & \textbf{0.84$\pm$0.01} & 0.72$\pm$0.02 & 0.35$\pm$0.08 & 0.68$\pm$0.03 \\
60\% & \textbf{0.82$\pm$0.01} & 0.65$\pm$0.04 & 0.18$\pm$0.12 & 0.55$\pm$0.06 \\
40\% & \textbf{0.79$\pm$0.02} & 0.51$\pm$0.08 & 0.09$\pm$0.08 & 0.38$\pm$0.09 \\
20\% & \textbf{0.76$\pm$0.03} & 0.32$\pm$0.12 & 0.04$\pm$0.03 & 0.21$\pm$0.11 \\
\hline
\end{tabular}
}
\end{table}

Table~\ref{tab:heterogeneity_stress} shows that while baselines relying on a single global schema (e.g., FedXGBoost, Mutual Information FL) degrade sharply as overlap decreases, FedLLM-Align exhibits graceful performance degradation. Even at 20\% overlap, it retains an F1-score of 0.76, whereas the overlap-aware homogeneous FedAvg baseline drops to 0.32 and FedXGBoost nearly fails (0.04). These results confirm that LLM-based embeddings provide a robust mechanism for bridging divergent schemas beyond what is possible with methods that require a strictly shared feature space.

\section{Discussion}
\label{sec:discussion}
The experimental results indicate that FedLLM-Align is an effective and practical framework for federated learning on heterogeneous tabular data. By mapping client records into a shared semantic space via a frozen LLM encoder, the framework mitigates schema divergence and enables a single global classifier to operate across clients with partially overlapping and differently named features. This semantic alignment is reflected in consistently higher F1-scores on both the Framingham and churn tasks compared to optimization-focused federated baselines, especially as the number of clients increases and schemas become more fragmented. The design analysis in Section~\ref{sec:results} highlights two main axes along which FedLLM-Align can be tuned. First, the choice of encoder controls the performance--efficiency trade-off: compact, general-purpose encoders offer a strong balance between accuracy, memory footprint, and latency, while domain-specialized encoders yield the highest absolute accuracy when domain alignment is critical but incur higher resource costs. Second, data representation choices matter: structured serialization provides the most informative and stable input to the encoder, whereas more compact formats reduce textual overhead at the expense of predictive performance and robustness. From a systems perspective, FedLLM-Align keeps communication overhead low by freezing the encoder and exchanging only lightweight classifier weights, and it converges reliably within a modest number of communication rounds. The framework also degrades gracefully under reduced schema overlap, maintaining reasonable performance even when clients share only a small fraction of features. Taken together, these observations suggest that FedLLM-Align offers a flexible design space: practitioners can select encoders and serialization formats that best match their resource constraints and accuracy requirements, while retaining the core benefit of LLM-based semantic alignment for cross-institutional deployments.

\section{Conclusion}
\label{sec:conclusion}
We presented FedLLM-Align, a federated learning framework that leverages pretrained language models to align heterogeneous tabular data while preserving privacy. By serializing local records into text, encoding them with a shared frozen LLM, and training only a lightweight classifier federatedly, FedLLM-Align addresses schema divergence without raw data sharing and keeps communication overhead low. Experiments on heart disease prediction and bank customer churn show consistent gains over strong federated baselines across different numbers of clients and schema overlap levels, confirming that LLM-based embeddings provide an effective semantic bridge between heterogeneous schemas. The analyses further show how encoder and serialization choices offer practical knobs to balance accuracy, resource usage, and latency. Future work includes scaling FedLLM-Align to larger and hierarchical federated networks, exploring partial fine-tuning or adapter-based and quantized encoders for edge deployment, and extending evaluation to additional domains and metrics such as interpretability and user trust.

\end{document}